%% file: ijcai26.tex
\documentclass{article}

\usepackage{ijcai26}

\usepackage{times}
\usepackage{soul}
\usepackage{url}
\usepackage[utf8]{inputenc}
\usepackage[small]{caption}
\usepackage{graphicx}
\usepackage{amsmath}
\usepackage{amsthm}
\usepackage{booktabs}
\usepackage{algorithm}

\usepackage[switch]{lineno}
\usepackage{xspace}
\usepackage{multirow}
\usepackage{makecell}
\usepackage{xspace}
\usepackage{amsmath,mathtools}
\usepackage{color}
\usepackage{xcolor}
\usepackage{bbding}
\usepackage{pifont}
\usepackage{wasysym}
\usepackage{amssymb}
\usepackage{diagbox}
\usepackage{amsthm}
\usepackage{newfloat}
\usepackage{listings}
\usepackage{algpseudocode}

\definecolor{todocolor}{rgb}{0.9,0.1,0.1}
\definecolor{hycolor}{rgb}{0.7,0.7,0.3}

\title{SAFformer: Improving Spiking Transformer via Active Predictive Filtering }

\author{
Zequan Xie$^1$
\and
Weiming Zeng$^2$
\and
Yunhua Chen$^{1}$\thanks{Corresponding authors.}
\and
Sichang Lin$^1$
\and
Tongyang Chen$^{1}$\And
Jinsheng Xiao$^3 $\\
\affiliations
$^1$ School of Computer Science and Technology, Guangdong University of Technology, Guangzhou, China\\
$^2$Faculty of Science, Hong Kong Baptist University, China\\
$^3$School of Electronic Information, Wuhan University, China\\
\emails
2112405284@mail2.gdut.edu.cn,
23267062@life.hkbu.edu.hk,
yhchen@gdut.edu.cn,
fatc811@163.com,
chentongyang@mails.gdut.edu.cn,
xiaojs@whu.edu.cn
}

\begin{document}

\maketitle

\begin{abstract}
Spiking Neural Networks (SNNs) offer notable advantages in biological plausibility and energy efficiency, making them promising candidates for building low-power Transformers. However, existing Spiking Transformers largely adhere to a \textit{passive reactive} paradigm, which struggles to focus on task-relevant information and incurs substantial computational overhead when processing redundant visual data. To overcome this fundamental yet underexplored limitation, we propose SAFformer, a novel Spiking Transformer architecture based on an \textit{active predictive filtering} paradigm. Inspired by the brain’s predictive coding mechanism, SAFformer actively suppresses predictable signals and focuses on salient visual features. Extensive experiments show that SAFformer establishes new state-of-the-art performance on CIFAR-10/100 and CIFAR10-DVS. Remarkably, on ImageNet-1K, it achieves 80.44\% Top-1 accuracy with only 26.58M parameters and an energy consumption of 5.88 mJ, demonstrating an exceptional balance between accuracy and efficiency. 
\end{abstract}

\section{Introduction}


Spiking neural networks (SNNs) are regarded as the third generation neural network models\cite{maass1997networks}. Their event-driven and sparse computing characteristics have demonstrated great potential in the field of artificial intelligence, especially in the small-scale edge computing scenarios\cite{guo2022real}. Unlike artificial neural networks (ANNs) that perform intensive computations at each inference step, SNNs transmit information asynchronously, with spiking neurons consuming energy only when receiving or emitting spikes\cite{eshraghian2023training}. In recent years, the integration of SNNs with the powerful Transformer architecture has shown great promise, paving the way for a new generation of visual models that achieve high performance with improved energy efficiency.

Despite this promise, a fundamental limitation persists in the design of existing Spiking Transformers. Whether in early direct-conversion attention mechanisms or later linear-complexity variants like QK-Attention\cite{zhang2024qkformer}, the information selection process universally follows a "Passive Reactive" paradigm. This model involves indiscriminate and costly processing of all information, followed by a post-hoc filtering step that relies solely on the fixed threshold of spiking neurons\cite{guo2025spiking,chen2024high}. Crucially, it lacks the ability to dynamically adjust its computational scope based on the input global context. Consequently, when faced with simple images containing large amounts of redundant background, the model still wastes significant computational resources on irrelevant regions. This paradigm runs counter to the highly efficient information processing mechanisms of the biological brain.
\begin{figure}[!t]
    \includegraphics[width=\linewidth]{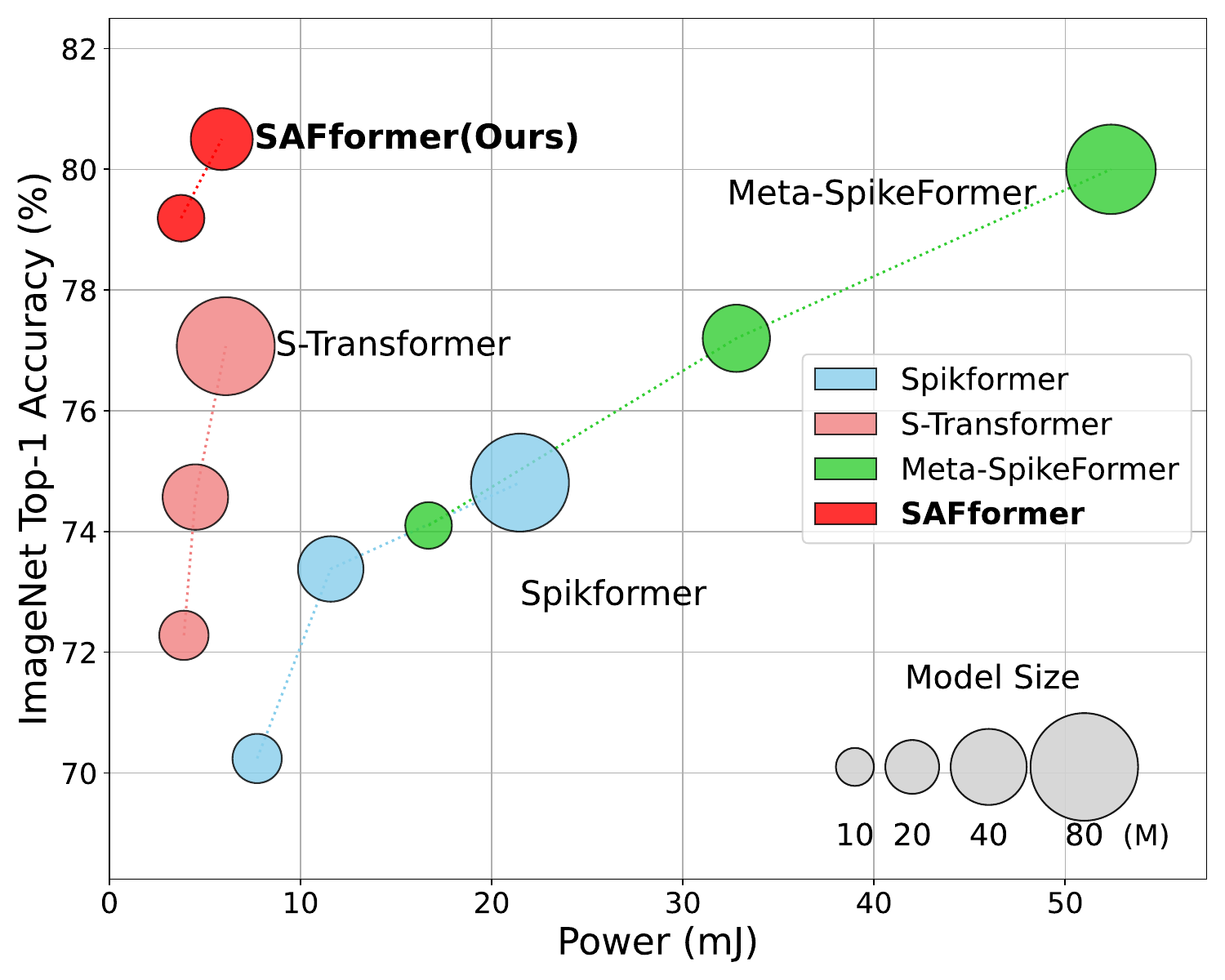}
    \caption{Comparison of our SAFformer with other Spiking Transformers on ImageNet-1k with an input resolution of 224 $\times$ 224.}
    \label{avm}
\end{figure}

In contrast, the predictive filtering mechanism inherent in the visual system enables the brain to utilize top-down feedback signals to modulate the activity of lower-level neurons\cite{crassidis1997predictive,rao1999predictive}. It actively suppresses predictable, redundant information and only forwards significant prediction errors for subsequent processing, thereby achieving extreme energy efficiency. Specifically, higher-level cortical areas (e.g. V4, IT) form a holistic prediction of the scene based on prior experience and send these predictive signals down to lower-level areas (e.g. V1, V2)\cite{friston2009free}. The lower-level cortex then compares this prediction with the actual visual input, and only the significant discrepancies are propagated upward as feedback signals, as illustrated in Fig. \ref{framework}(b).

Beyond the self-attention module, the MLP layer in the Transformer architecture, as an important part of network feature enhancement, is also a centralized area for computation and parameters. Vanilla Spiking MLPs, typically direct conversions of their ANN counterparts, suffer from two major flaws. First, they are spatially blind. Their 1$\times$1 convolutional structure can only mix information along the channel dimension\cite{wang2024multi} and is entirely incapable of perceiving or utilizing the local spatial context of tokens. Second, their feature transformation is static and homogeneous, applying an identical, unconditional transformation to all spatial locations\cite{chen2026c3net}. This inefficient processing model contradicts the principle of the visual cortex, which adjusts its processing intensity based on predictive information.

To systematically address these challenges, we draw inspiration from the brain's active prediction mechanism to propose SAFformer, a novel Spiking Transformer architecture that operates on the principle of Active Predictive Filtering. By incorporating two synergistic, bio-inspired innovations, SAFformer shifts the paradigm from passive reaction to active prediction. Our contributions are as follows:

 1) We propose the Spiking Active Predictive Filtering Attention(SAF), a new sparse spiking self-attention mechanism, as shown in Fig. \ref{SAF}. It introduces a lightweight Spiking Sparsity Guidance Module (SGM) to actively predict the saliency distribution of the input, thereby prospectively focusing attention resources on key spatial locations and realizing active predictive filtering. 

 2) We propose the Spiking Multi-scale Adaptive Gated (SMAG) network to replace the vanilla SMLP. SMAG perceives local spatial context through a decoupled, multi-scale gating mechanism, achieving more refined extraction and selection of predictive features while significantly reducing the number of parameters.
    
 3) Comprehensive experiments demonstrate that our architecture outperforms or matches SOTA Spiking Transformers on various datasets, with an excellent balance of efficiency and performance, as shown in Fig. \ref{avm}. Notably, we achieve SOTA accuracies of 97.50\% on CIFAR-10 and 83.38\% on CIFAR-100, setting a new benchmark for the SNN field.\footnote{The full version is available at https://arxiv.org/abs/2605.08270.}

\input{relatedwork}

\input{method}
\input{experiments}

\input{discussion}
\section*{Acknowledgments}

This work was supported by the Natural Science Foundation of Guangdong Province, China (No.2025A1515012243) and the Guangdong S\&T programme, China (No.2025A1010010002).

\bibliographystyle{named}
\bibliography{ijcai26}
\appendix
\input{supple}

\end{document}

%% file: relatedwork.tex
\section{Related Work}
\begin{figure*}[!h]
    \centering
    \includegraphics[width=1.0\linewidth]{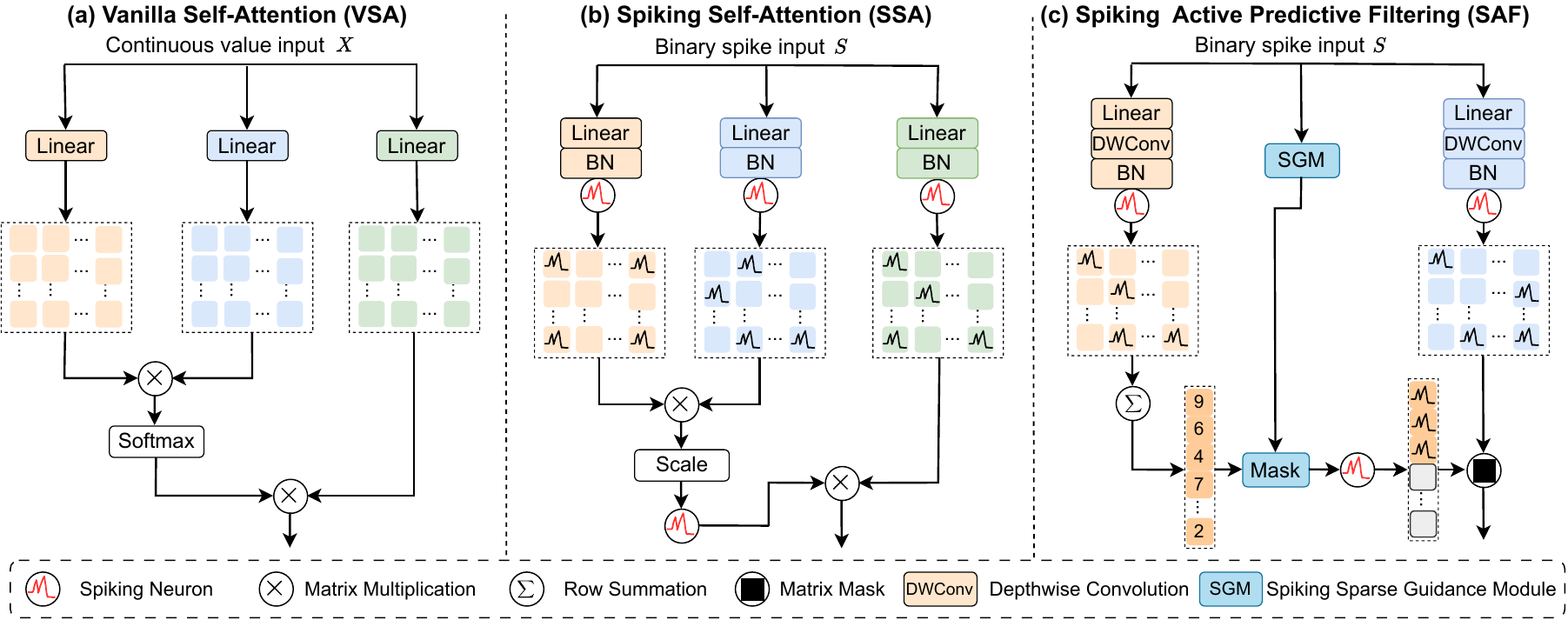}
    \caption{  Comparison of three self-attention paradigms. (a)VSA employs floating-point matrix multiplication to evaluate the spatial correlations between Q and K with quadratic computational complexity. (b)SSA transforms the floating-point operations in VSA into spiking computations, maintaining the same complexity as VSA. (c)SAF introduces a predictive filtering paradigm. It maintains linear complexity by using SGM to predict and sparsify significance scores from Q, which then act upon K element-wise.}
    \label{SAF}
\end{figure*}
\subsection{Spiking Neural Networks}

Spiking Neural Networks (SNNs) are widely recognized as the third generation of neural network models, with their design inspired by the information processing mechanisms of the biological brain. Unlike traditional ANNs, SNNs communicate by transmitting binary spikes, mimicking the firing of biological neurons\cite{duan2023neurozoom}. When the membrane potential of a spiking neuron exceeds a certain threshold, it fires a spike, represented as '1'; otherwise, it remains silent, represented as '0'\cite{yao2024spike}. This event-driven mechanism facilitates sparse synaptic operations and avoids costly multiply-accumulate (MAC) operations\cite{chen2026sqkformer}, thereby significantly enhancing the energy efficiency of these models. SNNs are primarily trained using two methods: ANN-to-SNN conversion and direct training with surrogate gradients\cite{di2024ec}. ANN-to-SNN conversion involves training an ANN and then converting it to a corresponding quantized SNN version. Direct training, on the other hand, employs a surrogate function to approximate the spike firing process during backpropagation\cite{yao2024spike}. The effectiveness of direct training in handling temporal data has made it a preferred choice in SNN research.

However, conventional direct training methods require substantial GPU memory due to the simulation of multiple timesteps\cite{luo2024integer,qiu2025quantized}. To address the challenge of long training times, Yao introduced a novel direct training method called Spike Firing Approximation (SFA)\cite{yao2025scaling}. This method uses quantized membrane potentials as activation values during the training phase and converts these values into spike sequences during inference, thereby reducing training duration. It has been demonstrated that SFA can reduce training time by at least 4.1$\times$.

\subsection{Spiking Vision Transformers}
The Transformer, with its powerful sequence modeling capabilities and self-attention mechanism, has become a mainstream backbone in the field of computer vision\cite{dosovitskiy2020image}. However, its quadratic computational complexity and significant memory footprint limit its application in power-sensitive and resource-constrained scenarios\cite{xiao2026stkps}. Combining the powerful capabilities of the Transformer with the low-power and event-driven characteristics of SNNs has become an active research direction.

Spikformer\cite{zhou2023spikformer} pioneered spiking self-attention computation, establishing the first spiking ViT. Spikingformer\cite{zhou2026spikingformer} further enhanced the SNN compatibility of the model by designing a pre-activation residual connection. To achieve a fully spike-driven architecture, Spike-driven Transformer\cite{yao2023spike} implemented the Hadamard product in its self-attention module, becoming the first directly trained, fully spike-driven transformer model. To reduce temporal complexity, QKFormer\cite{zhang2024qkformer} proposed using only Q and K for self-attention computation, becoming the first spiking Transformer with a linear-complexity self-attention module. Additionally, OST\cite{song2024one} compresses the time domain into 1 to enhance the training speed. To fully address the issue of over-allocating attention, SpiLiFormer\cite{zheng2025spiliformer} introduced a lateral inhibition mechanism to simulate the inhibitory-excitation interaction process.

%% file: method.tex
\section{Methods}


\begin{figure*}[htb]
    \centering
    \includegraphics[width=1.0\linewidth]{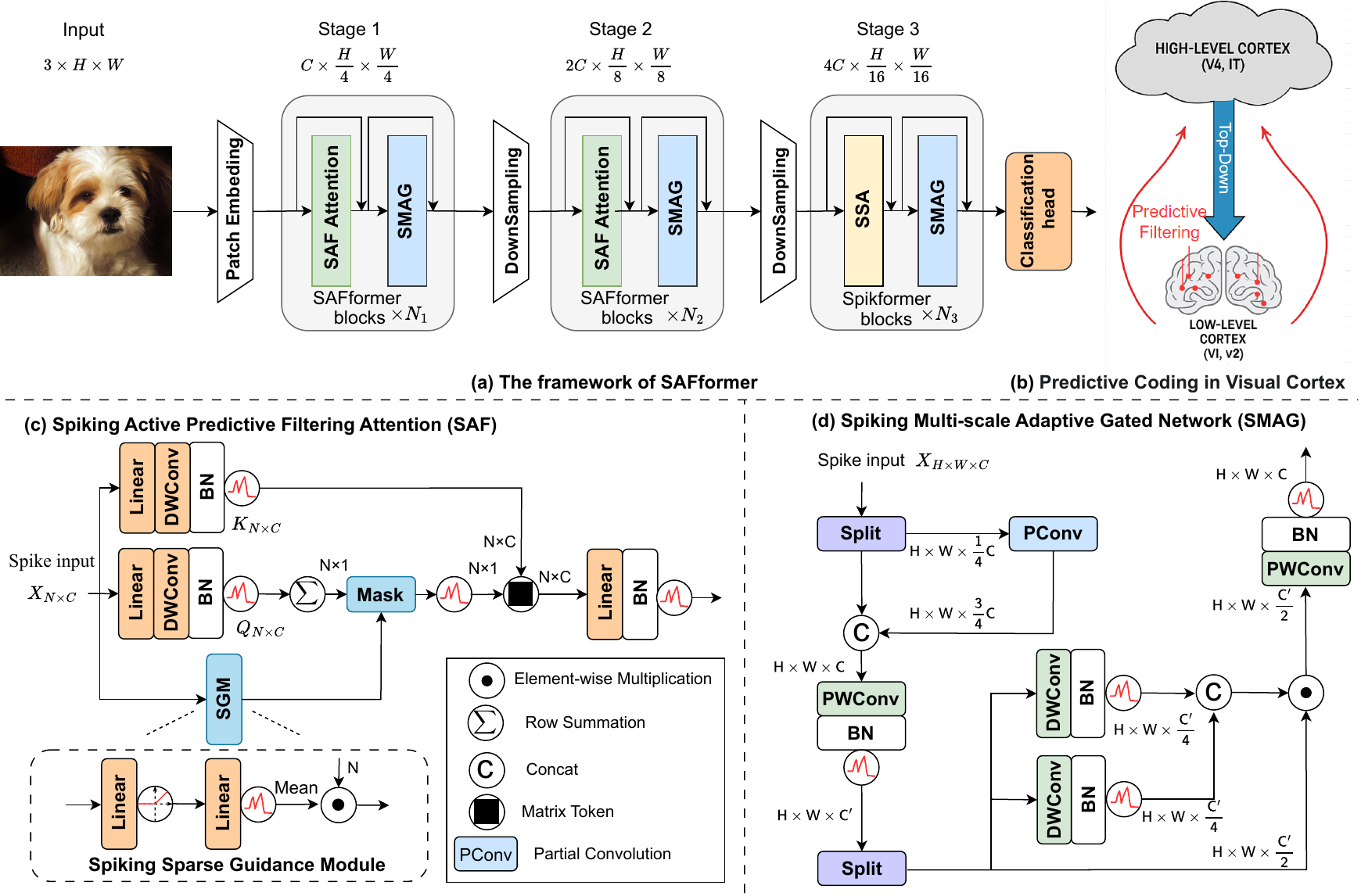}
    \caption{  (a) The architecture of SAFformer. We propose the SAFformer block, which consists of SAF Attention and SMAG. (b) An illustration of the predictive coding mechanism in visual cortex. (c) and (d) illustrate the workflows of SAF Attention and SMAG.}
    \label{framework}
\end{figure*}

In this section, we first introduce the spiking neuron model we used. Next, we present the SAFformer architecture, which includes a novel spiking active predictive filtering attention mechanism and a multi-scale gated feedforward network.
\subsection{Preliminary}

In this paper, we use the widely adopted LIF model \cite{guo2024ternary}, which can be described by the following formula:
\begin{align}
    U[t] &= \beta H[t - 1] + X[t], \label{one} \\
    S[t] &= \operatorname{Hea}(U[t] - V_{\text{th}}), \label{two} \\
    H[t] &= 
    \begin{cases}
    U[t] \cdot (1 - S[t]) + V_{\text{reset}} \cdot S[t], & \text{hard reset}, \\
    U[t] - V_{\text{th}} \cdot S[t], & \text{soft reset},
    \end{cases} 
\end{align}
where $t$ is the time step, $U[t]$ is the membrane potential, the spatial input $X[t]$ is extracted from the original spike sequence via Conv or Linear layers, and the temporal input $\beta H[t - 1]$ comes from the decayed membrane potential of the previous time step, $\beta$ is the leak factor. $S[t]$ is the output spike, and $H[t]$ is the membrane potential after firing. $\operatorname{Hea(\cdot)}$ is the Heaviside step function, where $\operatorname{Hea(x)} = 1$ if $x \geq 0$. When $U[t]$ exceeds the threshold $V_{\text{th}}$, the spiking neuron fires a spike, and the membrane potential is reset to $V_{\text{reset}}$.

\subsection{Overall Architecture}

The overall architecture of SAFformer is shown in Fig. \ref{framework}(a), which consists of three stages. The number of blocks (SAFformer blocks or Spikformer blocks) in each stage is represented as $N_1$, $N_2$, and $N_3$, respectively. The input data is represented as $I \in \mathbb{R}^{T \times C_{in} \times H \times W}$, where $C_{in}=3$ for static image data and $C_{in}=2$ for neuromorphic data. In the first stage, a patch size of $4 \times 4$ is utilized. 
A Patch Embedding module converts the input into a spike sequence and maps the feature of each patch to an initial channel dimension $C$, reducing the number of tokens to $\frac{H}{4} \times \frac{W}{4}$.The transformed features are then processed through SAF and SMAG. The processing mechanism in the second stage is similar, but with a patch size of $2 \times 2$, an expanded channel dimension of $2C$, and a reduced token count of $\frac{H}{8} \times \frac{W}{8}$. The third stage is the deep stage, where we continue to use a $2 \times 2$ patch size, expand the channel dimension to $4C$, and reduce the token count to $\frac{H}{16} \times \frac{W}{16}$.

\subsection{Spiking Active Predictive Filtering(SAF)}

We propose Spiking Active Predictive Filtering Attention (SAF), whose detailed structure is shown in Fig. \ref{framework}(c). The core idea of SAF lies in that it no longer calculates the pairwise relationships between tokens. Instead, it employs a lightweight guidance module to predict the most information-dense regions and dynamically focuses computational resources on them. Considering that the intrinsic low-pass filtering characteristics of spiking neurons can attenuate high-frequency signals\cite{meta2024spike,fang2025spiking}, we introduce a depthwise separable convolution (DWConv) into the generation paths of the spike matrices $Q$ and $K$. This is to better preserve local details crucial in the shallow stages of the network. For a given input spike sequence $X \in \mathbb{R}^{N \times C}$, where $N$ is the number of tokens and $C$ is the channel dimension, the generation is formulated as:
\begin{algorithm}[htb]
\caption{Spiking Sparsity Guidance Module (SGM)}
\label{alg:sgm}
\begin{algorithmic}[1]
\Statex \textbf{Input:} $X \in \mathbb{R}^{T\times B\times C\times N}$ \hfill $\triangleright$ Input spike sequence
\Statex \textbf{Output:} Dynamic $k$ \hfill $\triangleright$ The number of tokens to keep
\State $F \leftarrow \text{Linear}_1(X)$ \hfill $\triangleright$ Project to hidden space
\State $F \leftarrow \text{ReLU}(F)$ \hfill
\State $G_{raw} \leftarrow \text{Linear}_2(F)$ \hfill $\triangleright$ Generate guidance scores
\State $G_{raw} \leftarrow \text{SN}(G_{raw})$ \hfill $\triangleright$ Shape: $T \times B \times 1 \times N$
\State $G_{flat} \leftarrow \text{Flatten}(G_{raw})$ \hfill $\triangleright$ Generate a vector of shape
\State $p \leftarrow \frac{1}{TBN}\sum\limits_{t,b,n} G_{\mathrm{flat}}[i]$ \hfill $\triangleright$ Global sparsity ratio
\State Dynamic $k \leftarrow \lfloor N\cdot p \rfloor$ \hfill $\triangleright$ Global scaling factor,where N is the token count
\State \Return Dynamic $k$
\end{algorithmic}
\end{algorithm}
\begin{align}
       {Q}&=\mathrm{SN}\left(\mathrm{BN}\left(\mathrm{DWConv}\left({X}{W}_{Q}\right)\right)\right), \\{K}&=\mathrm{SN}\left(\mathrm{BN}\left(\mathrm{DWConv}\left({X}{W}_{K}\right)\right)\right),
\end{align}
where $W_Q$ and $W_K$ represent the linear projections of the input spike sequence $X$, and $\text{SN}(\cdot)$ is the spiking activation function. Then, we obtain the initial saliency score for each token by summing over the channel dimension of $Q$:
\begin{align}
A_{t} = \sum_{i=0}^{C} Q_{i,j}
\end{align}
To actively filter redundant information and adaptively adjust the computational load based on content complexity, we develop a lightweight and learnable module, named Spiking Sparsity Guidance Module (SGM). It dynamically predict a suitable attention matrix sparsity level, $k$, based on the global information of the input. In Algorithm~\ref{alg:sgm}, we detail the full process of generatingdynamic $k$ using SGM.

With the dynamic $k$ generated by SGM, a masking operation is performed on $A_t$ to create a binary mask matrix $M_k$:
\begin{align}
    (M_{k})_{i} = 
\begin{cases} 
1, & \text{if } (A_{t})_i \in \text{Topk}(A_{t}, k) \\ 
0, & \text{otherwise} 
\end{cases}
\end{align}
where $\text{Top-K}(\cdot, k)$ denotes the selection of the $k$ largest elements from the vector.

Finally, the mask $M_k$ is applied to the saliency scores, and the result is multiplied element-wise with $K$ in a Hadamard product:
\begin{align}
       \mathrm{SAF}(Q,K) = K \cdot\mathrm{SN}(M_{k} \odot A_{t}) 
\end{align}
\subsubsection{Theoretical Analysis}
To formally analyze the fundamental differences in information processing across attention mechanisms, we introduce information flow graph\cite{ahlswede2000network} as a unified analytical framework.

An information flow graph is a weighted directed graph $G=(V, E, W)$, where the vertex set $V = \{v_1, \dots, v_N\}$ represents the input tokens, a directed edge $(v_j, v_i) \in E$ signifies an information flow from token $j$ to token $i$, and the weight function $W: E \to \mathbb{R}$ assigns a weight $w_{ij}$ to each edge.
The output representation $y_i$ for token $i$ is a weighted combination of information from its inbound neighbors: 
\begin{align}
    y_i = f\left(\sum_{v_j \in \text{In}(v_i)} w_{ij} \cdot \text{Info}(v_j)\right)
\end{align}
where $\text{In}(v_i)$ is the set of inbound neighbors of $v_i$, $\text{Info}(v_j)$ is the information carried by token $v_j$, and $f(\cdot)$ is a non-linear activation function.

\paragraph{Proposition 1.} \textit{The computational paradigm of SSA is equivalent to performing a relational aggregation operation on a dense, fully-connected graph.}

\textit{Proof.} The output $y_i$ of SSA is determined by the equation:
\begin{equation}
    y_i = f\left(\sum_{j=1}^{N} \langle Q_i, K_j \rangle \cdot V_j\right)
\end{equation}
where $Q_i, K_j, V_j$ are the Query,Key,Value vectors derived from tokens $i$ and $j$. The summation $\sum_{j=1}^{N}$ iterates over all input tokens $v_j$ for any given output $y_i$. This implies that for any pair of vertices $(v_i, v_j)$, an information flow path, i.e., a directed edge $(v_j, v_i)$, exists. Therefore, the graph is a complete graph where the edge set is $E = V \times V$.

And the weight of an edge $(v_j, v_i)$, denoted $w_i,j$, is given by the inner product $\langle Q_i, K_j \rangle$. This weight is a binary relational function $R(Q_i, K_j)$ that quantifies the pairwise relation between vertices $v_i$ and $v_j$. The final output $y_i$ is a weighted sum of the Value information $v_j$ from all vertices in the graph. This process, which gathers information from the entire graph, is a quintessential aggregation operation.
\paragraph{Proposition 2.} \textit{The computational paradigm of SAF is equivalent to performing a node-wise filtering operation on a dynamically sparse subgraph.}

\textit{Proof.} The output $y_i$ of SAF can be represented as:
\begin{equation}
    y_i = f\left((M_k)_i \cdot \sum_{c=1}^{C} Q_{c,i} \cdot K_i\right)
\end{equation}
where $q_i, k_i$ are derived from token $i$, and $(M_k)_i$ is a dynamic mask.
The computation of $y_i$ depends solely on its own derivatives, implying that no cross-vertex edges exist. The information flow graph thus consists only of self-loops. 

However, not all vertices participate in the computation. A global selection function $\Psi(X, k)$, realized by the SGM, first identifies an active vertex subset $V_{\text{active}} \subseteq V$.
    \begin{equation}
        V_{\text{active}} = \{v_i \in V \mid i \in \text{Topk}(\{ \sum_c Q_{c,j} \}_{j=1}^N, k)\}
    \end{equation}
where $|V_{\text{active}}| = k$. The effective computation graph $G'$ is thus a dynamic sparse subgraph consisting only of the vertices in $V_{\text{active}}$.

For any active vertex $v_i \in V_{\text{active}}$, the weight of its self-loop, $w_{ii}$, is determined by a unary saliency function $S(v_i)$ which is computed as $\sum_{c=1}^{C} Q_{c,i}$. This weight is a function of the vertex's intrinsic properties, not its relation to others. The output $y_i$ is a scaled version of the vertex's own information $K_i$, which constitutes a filtering operation. Since the selection of the subgraph is dynamically determined by the global input $X$, the process is an active adaptive filtering. 

SAF facilitates a paradigm shift in spiking self-attention, from relational aggregation to dynamic adaptive filtering, offering a new perspective for building efficient and powerful attention mechanisms in Spiking Neural Networks.

\subsection{Spiking Multi-scale Adaptive Gated Network(SMAG)}
Spiking MLP is a key part of the Spiking Transformer for improving feature representation. However, the commonly used SMLP with wide intermediate channels has the disadvantages of excessive parameter quantity and spatial blind viewing\cite{qiu2024gated}, which violates the principle of active predictive filtering that we pursue.

Therefore, we propose a novel feedforward network, the Spiking Multi-scale Adaptive Gated Network (SMAG), to replace the vanilla SMLP, as illustrated in Fig. \ref{framework}(d). SMAG performs feature transformation in a refine-and-gate paradigm, enhancing performance while reducing the burden of processing redundant information. Specifically, we first introduce a partial convolution\cite{chen2023run} to a fraction of the channels of the input spike features $X \in \mathbb{R}^{T \times B \times C \times H \times W}$. This operation injects valuable local spatial priors into the feature maps at a minimal computational cost. Subsequently, the features are decoupled into three independent functional paths as follows:
\begin{align}
    &X^{\prime}
    =\mathrm{SN}\left(\mathrm{BN}\left(W_{1} \mathrm{PConv}(X)\right)\right), \\
    &\left[X_{1}^{\prime},
    X_{2}^{\prime},X_{3}^{\prime}\right]=X^{\prime}.
\end{align}

where $\mathrm{PConv}(\cdot)$ is the partial convolution operation, and $[, ]$ denotes the channel-wise splitting, and $W_{1}$ represents a pointwise convolution. PConv effectively introduces local context, which provides a foundation for generating more context-aware gating signals by initially filtering and strengthening a subset of features. Then, $X_{1}^{\prime}$ and $X_{2}^{\prime}$ serve as gating signals, while $X_{3}^{\prime}$ acts as the main path. A multi-scale dynamic gating mechanism is introduced to capture spatial patterns at multiple scales:
\begin{align}
    &G_{1}^{\prime}  =\mathrm{SN}\left(\mathrm{BN}\left(\mathrm{DWConv}_{3\times3}\left(X_{1}^{\prime}\right)\right)\right), \\
     &G_{2}^{\prime}  =\mathrm{SN}\left(\mathrm{BN}\left(\mathrm{DWConv}_{7\times7}\left(X_{2}^{\prime}\right)\right)\right), \\
    &X_\mathrm{gated}^{\prime}  = \mathrm{Concat}(G_{1},G_{2})\odot X_{3}^{\prime}.
\end{align}
where $\mathrm{DWConv}_{3\times3}$ and $\mathrm{DWConv}_{7\times7}$ are depthwise convolutions with kernel sizes of 3 and 7, respectively. They learn spatial filters independently for each channel, significantly reducing parameters and computational complexity compared to standard convolutions, while effectively capturing channel-specific spatial patterns. The parallel design of 3x3 and 7x7 kernels enables the gating signals to perceive both fine-grained details and broader regional context simultaneously. Finally, through the element-wise gating multiplication, active suppression and enhancement of the content stream are achieved.

Theoretically, SMAG has significantly fewer parameters than vanilla SMLP. Moreover, compared to the two identical linear transformations in SMLP, the dynamic gating mechanism of SMAG endows the network with the ability to adaptively select and modulate feature channels based on the input content. This allows the model to focus more on critical information and suppress redundancy, thereby improving the discriminability of features.

%% file: experiments.tex
\section{Experiments \label{experiments}}

\begin{table*}[htb]
\centering 
\begin{tabular}{lllcccc}
\toprule[1pt]\multirow{2}{*}{Methods}                                                 & \multirow{2}{*}{Type} & \multirow{2}{*}{Architecture} & \multicolumn{1}{l}{Param} & \multicolumn{1}{l}{Time} & \multicolumn{1}{l}{Power} & Top-1          \\
&                       &                               & (M)                       & Step                     & (mJ)                      & Acc(\%)        \\ \hline\multirow{2}{*}{DeiT\cite{touvron2021training}}         & ANN                   & DeiT-S                        & 22.00                     & 1                        & 21.20                     & 79.80          \\
& ANN                   & DeiT-B                        & 86.59                     & 1                        & 80.50                     & \textbf{81.80} \\ \hline
MST\cite{wang2023masked}                                & A2S                   & Swin Transformer-T            & 28.50                     & 512                      & -                         & 78.51          \\
STA\cite{hwang2024spikedattention}                      & A2S                   & ViT-B/32                      & 86.00                     & 32                       & -                         & \textbf{78.72} \\ \hline\multirow{2}{*}{Spikformer\cite{zhou2023spikformer}}    & SNN                   & Spikformer-8-384              & 16.81                     & 4$\times$1               & 7.73                      & 70.24          \\
& SNN                   & Spikformer-8-512              & 29.68                     & 4$\times$1               & 11.58                     & 73.38          \\ \cline{2-7} \multirow{2}{*}{S-Transformer v2\cite{yao2024spike}}    & SNN                   & S-Transformer v2-8-384        & 15.10                     & 4$\times$1               & 16.70                     & 74.10          \\
& SNN                   & S-Transformer v2-8-512        & 31.30                     & 4$\times$1               & 32.80                     & 77.20          \\ \cline{2-7} \multirow{2}{*}{Max-Former\cite{fang2025spiking}}                                              & SNN                   & Max-10-384                    & 16.23                     & 4$\times$1               & 4.89                      & 77.82          \\
& SNN                   & Max-10-512                    & 28.65                     & 4$\times$1               & 7.49                      & 79.86  \\\cline{2-7} \multirow{2}{*}{MSViT\cite{hua2025msvit}}                                              & SNN                   & MSViT-10-384                    & 17.69                     & 4$\times$1               & 16.65                      & 80.09          \\
& SNN                   & MSViT-10-512                    & 30.23                     & 4$\times$1               & 24.74                     & 82.96        \\ \cline{2-7} \multirow{2}{*}{E-SpikeFormer\cite{yao2025scaling}}                                           & SNN                   & E-SpikeFormer-S               & 10.00                     & 1$\times$4               & 3.00                      & 78.50          \\
& SNN                   & E-SpikeFormer-M               & 19.00                     & 1$\times$4               & 5.90                      & 79.80          \\ \cline{2-7} \multirow{2}{*}{QSD-Transformer\cite{qiu2025quantized}} & SNN                   & S-Transformer v2-T$^\dagger$  & 1.80                      & 1$\times$4               & 2.50                      & 77.50          \\
& SNN                   & S-Transformer v2-M$^\dagger$  & 3.90                      & 1$\times$4               & 5.70                      & 78.90          \\ \hline\multirow{2}{*}{\textbf{SAFformer}}                                      & SNN                   & SAFformer-10-384              & 15.12                     & 1$\times$4               & 3.75                      & 79.19          \\
& SNN                   & SAFformer-10-512              & 26.58                     & 1$\times$4               & 5.88                      & \textbf{80.44$\pm$ 0.24} \\ 
\bottomrule[1pt]\end{tabular}
\caption{Evaluation on ImageNet. All models default to an input size of 224 $\times$ 224. We have reformatted the time steps of all directly trained SNNs into the $T\times D$ format, where $T$ is the number of time steps, and $D$ is the upper limit of integer activation during training. $D\geq 1$ signifies the use of the SFA direct training method. The $\dagger$ symbol indicates the use of knowledge distillation.}
\label{imagex}
\end{table*}

\begin{table*}[htb]
\centering
\begin{tabular}{lcccccccc}
\toprule[1pt]\multirow{2}{*}{\textbf{Method}}                     & \multicolumn{2}{c}{CIFAR10}                         & \multicolumn{2}{c}{CIFAR100}                        & \multicolumn{2}{c}{CIFAR10-DVS}                     & \multicolumn{2}{c}{DVS128}                          \\ \cline{2-9} 
                                                     & \multicolumn{1}{r}{Param} & \multicolumn{1}{r}{Acc} & \multicolumn{1}{r}{Param} & \multicolumn{1}{r}{Acc} & \multicolumn{1}{r}{Param} & \multicolumn{1}{r}{Acc} & \multicolumn{1}{r}{Param} & \multicolumn{1}{r}{Acc} \\ \hline
FSTA-SNN\cite{yu2025fsta}          & 11.30                     & 96.52                   & 11.30                     & 80.42                   & 11.20                     & 82.7                    & -                         & -                       \\
STAA-SNN\cite{zhang2025staa}       & -                         & 97.14                   & -                         & 82.05                   & -                         & 82.1                    & -                         & 98.6                    \\ \hline
Spikformer\cite{zhou2023spikformer}& 9.32                      & 95.51                   & 9.32                      & 78.21                   & 2.57                      & 80.9                    & 2.57                      & 98.3                    \\
S-Transformer\cite{yao2023spike}   & 10.28                     & 95.60                   & 10.28                     & 78.40                   & 2.57                      & 80.0                    & 2.57                      & \textbf{99.3}                    \\
QKFormer\cite{zhang2024qkformer}    & 6.74                      & 96.18                   & 6.74                      & 81.15                   & 1.50                      & 84.0                    & 1.50                      & 98.6                    \\
MSViT\cite{hua2025msvit}                                                & 7.59                      & 96.53                   & 7.59                      & 81.98                   & 1.67                      & 84.0                    & 1.67                      & 98.8                    \\
Max-Former\cite{fang2025spiking}                                           & 6.57                      & 97.04                   & 6.60                      & 82.65                   & 1.45                      & 84.2                    & 1.45                      & 98.6                    \\ \hline \textbf{SAFformer}                     & 6.31                      & \textbf{97.50}                   & 6.34                      & \textbf{83.38}                   & 1.50                      & \textbf{85.0}                    & 1.50                      & 99.0                    \\ \bottomrule[1pt]
\end{tabular}
\caption{Comparison on CIFAR10, CIFAR100, DVS128, and CIFAR10-DVS.}
\label{cifex}
\end{table*}
In this section, we conduct a comprehensive set of experiments to evaluate our proposed method, comparing it with other recent state-of-the-art architectures on several widely-used benchmarks. We evaluate our model on various datasets, including the static datasets including CIFAR-10/100\cite{krizhevsky2009learning} and ImageNet\cite{deng2009imagenet}, as well as neuromorphic datasets CIFAR10-DVS\cite{li2017cifar10} and DVS128Gesture\cite{amir2017low}.

\subsection{Results on ImageNet-1K Classification }
\paragraph{Experimental Setup on ImageNet-1K.}
The input image size is set to 224 $\times$ 224. For model training, we employ the SFA direct training method\cite{yao2025scaling} with the virtual timestep $D$ set to 4. We use the AdamW optimizer with a base learning rate of $6 \times 10^{-4}$, where the actual learning rate is calculated by multiplying the base rate by the batch size divided by 256. Training is conducted for 200 epochs with a batch size of 512, distributed across 6 NVIDIA RTX 3090 GPUs. Throughout the process, we apply standard data augmentation techniques, including RandAugment\cite{cubuk2020randaugment}, mixup, and random erasing. The number of blocks in the three stages is set to $\left \{ 1,2,7\right \} $, respectively.

\begin{table}
    \centering
    \begin{tabular}{cccc}
\toprule[1pt]
SAF      & SMAG  & Param(M) & Top-1 Acc(\%)   \\ \hline
\multicolumn{2}{c}{baseline}                          & 6.74                         & 79.79        \\ \hline
\checkmark &                           & 6.77                         & 80.42(+0.63) \\                     & \checkmark  & 6.32                         & 83.07(+3.28) \\
\checkmark & \checkmark  & 6.34                         & 83.38(+3.59) \\ \toprule[1pt]
\end{tabular}
    \caption{Ablation study of SAFformer on CIFAR100.}
    \label{aba}
    \vspace{-0.4cm}
\end{table}
\noindent\textbf{Primary Results on ImageNet-1K.}
The experimental results are shown in Table \ref{imagex}. Compared with the direct training method of vanilla time steps, our SAFformer (26.58M, 80.44\%) achieves a 2.24\% higher accuracy than S-Transformer v2 (31.30M, 77.20\%) while reducing the number of parameters by approximately 15.1\%. Compared with the direct training method of SFA, under the same virtual time step $D=4$, our model has a 0.64\% higher accuracy rate and similar energy consumption compared with E-SpikeFormer. Compared to the latest Max-Former (28.65M, 79.86\%), SAFformer achieves a 0.58\% performance improvement with fewer parameters and lower energy consumption. Without considering knowledge distillation, our model achieves comparable accuracy and efficiency to SOTA models in the SNN domain on ImageNet-1K.

\subsection{Results on CIFAR and Neuromorphic Datasets} \label{sec42}
\paragraph{CIFAR Classification.}
On CIFAR datasets, SAFformer is trained for 400 epochs with a batch size of 64, and the virtual timestep $D$ is set to 4. We use the AdamW optimizer with a learning rate of $1 \times 10^{-3}$. The SAFformer consists of 4 blocks, distributed across the three stages as $\left \{ 1,1,2\right \} $. The results are shown in Table \ref{cifex}. On CIFAR10, our model achieves 97.50\% accuracy with 6.31M parameters, surpassing all previous models including Max-Former (97.04\%) and STAA-SNN (97.14\%). On the more challenging CIFAR100 dataset, SAFformer achieves an accuracy of 83.38\%, which is also significantly higher than the previous SOTA models Max-Former (82.65\%) and STAA-SNN (82.05\%). 

\noindent\paragraph{Neuromorphic Classification.}
The neuromorphic datasets use a SAFformer with 1.50M parameters, configured with a $\left \{ 0,1,1\right \} $ block distribution across its three stages and a timestep of 16. The training process for DVS128Gesture involves 200 epochs, while for CIFAR10-DVS it is 106 epochs. On CIFAR10-DVS, our model achieves an accuracy of 85.0\%, significantly outperforming QKFormer (84.0\%) and Max-Former (84.2\%) to establish a new state-of-the-art. On the DVS128 Gesture recognition task, SAFformer achieves an accuracy of 99.0\%, demonstrating highly competitive performance.

\begin{figure}[htb]
    \includegraphics[width=\linewidth]{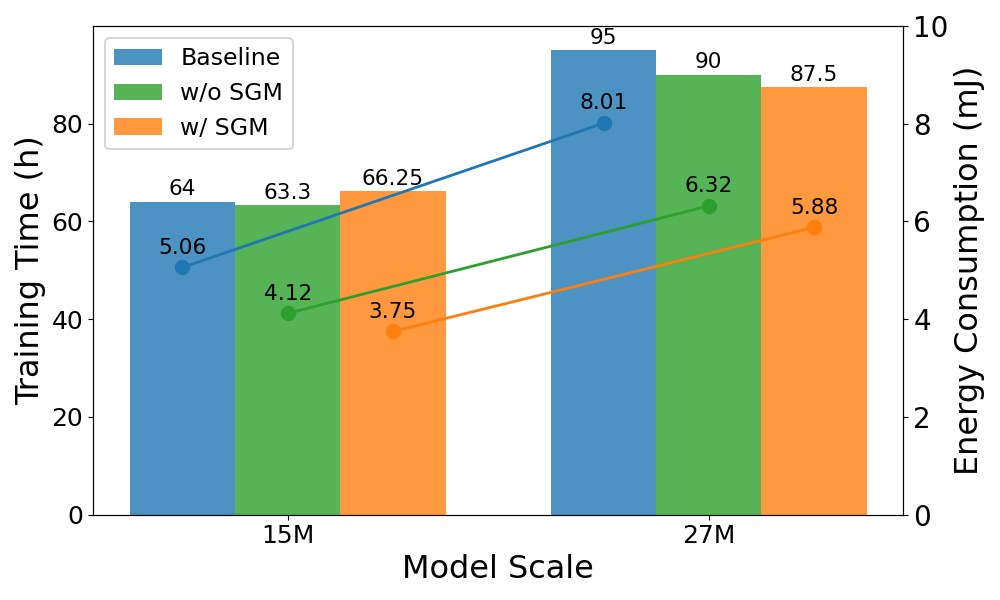}

    \caption{Ablation study on SGM, analyzing its impact on training time and energy consumption across different model scales.}
    \label{saftime}
\end{figure}
\vspace{-0.4cm}

\subsection{Ablation Study}

\begin{table}[htb]
\centering
\begin{tabular}{lcc}
\toprule[1pt]
Methods             & \multicolumn{1}{l}{Param(M)} & \multicolumn{1}{l}{Top-1 Acc(\%)} \\ \hline
w/o depthwise conv         & 6.77                         & 80.42                             \\ \hline
w/o partial conv          & 6.23                         & 82.51                             \\ \hline
w/o multi-scale    & 6.36                         & 83.15                             \\ \hline
\textbf{SMAG(ours)} & \textbf{6.34}                & \textbf{83.38}                    \\ \toprule[1pt]
\end{tabular}
\vspace{-0.2cm}
\caption{Ablation study on the components of the SMAG module.}
    \label{smagaba}
    \vspace{-0.4cm}
\end{table}

We conduct ablation studies on CIFAR100 to analyze the effectiveness of each module, with all experiments following the training details described in Section \ref{sec42} unless otherwise specified. The experimental results are presented in Table \ref{aba}. The results show that each component contributes positively to the final performance. Specifically, SAF has improved its accuracy by 0.63\% with almost no increase in the number of parameters by dynamically predict the tokens with the richest information. Furthermore, substituting the SMLP with our SMAG yields a substantial 3.28\% accuracy gain while simultaneously reducing the parameter count by 0.42M. The best performance is achieved when both SAF and SMAG modules are combined, demonstrating that they are effective and complementary.

\noindent\paragraph{{Effectiveness of the SGM Module in SAF.}}
The effectiveness of the SGM module is validated in Fig. \ref{saftime}. We use a SAFformer without any filtering mechanism as the baseline. The "w/o SGM" version removes SGM and uses a fixed Top-K ratio instead. For the 27M model, SGM reduces energy consumption by approximately 27\% compared to the baseline and accelerates training time by 8\%. Our SAFformer (15.12M, 3.75mJ,79.19\%) outperforms the Fixed Top-k baseline (15.08M,
4.12mJ, 78.71\%), which demonstrates SGM’s effective efficiency and accuracy. While the training time is slightly longer for the smaller 15M model due to relative overhead, the substantial energy savings in all scenarios confirm the superior efficiency of our active predictive approach.

\noindent\paragraph{Component Analysis of the SMAG Module.}
We analyzed the contribution of each key design in SMAG through a step-by-step ablation, with results in Table \ref{smagaba}. When we remove the multi-scale design of SMAG, the model's performance drops to 83.15\%. Further removing the partial convolution reduces accuracy to 82.51\%. Completely eliminating spatial awareness by replacing all depthwise convolutions with $1\times1$ convolutions causes the most significant performance degradation, with accuracy falling to 80.42\%. These results clearly demonstrate that the superiority of SMAG stems from the synergistic combination of its spatial awareness, local prior injection, and multi-scale gating. 
\vspace{-0.2cm}

%% file: discussion.tex
\section{Conclusion}
In this paper, we have delved into the inherent Passive Reactive limitation in existing Spiking Transformers. Drawing inspiration from the brain's predictive coding theory, we proposed SAFformer. It actively predicts and filters redundant information, guiding the model to focus its computational resources on salient features while preserving more high-frequency information. We introduced a new self-attention mechanism, SAF, which actively predicts global saliency via a lightweight guidance module to dynamically focus attention resources on key regions. Furthermore, we proposed SMAG to replace the vanilla SMLP. It leverages a decoupled, multi-scale gating mechanism to overcome homogeneous transformation and parameter redundancy. Experimental results demonstrate that SAFformer achieves state-of-the-art performance on CIFAR-10/100 and CIFAR10-DVS datasets while maintaining low energy consumption.

%% file: supple.tex
\section{Energy Consumption Calculation}
    This section elaborates in detail the methods we use to evaluate the energy consumption of the SAFformer model. When assessing energy consumption, we follow the most commonly used energy consumption evaluation method in the SNN field. We mainly focus on the energy consumption of key operations in the computing process (such as multiplication and accumulation operations MAC and accumulation operations AC), and ignore some overhead related to hardware manufacturing, data loading and storage. We assume that all operations are based on 32-bit floating-point precision and refer to widely used industry benchmarks to set energy consumption parameters. At the 45nm technology node, the energy consumption for MAC operation is approximately $E_{MAC} = 4.6 pJ$, while the energy consumption for AC operation is approximately $E_{AC} = 0.9 pJ$. For specific ANNs and SNNs, the energy consumption calculation formulas are as follows:
    \begin{align}
         E_{A N N}=&O^{2} \times C_{\text {in }} \times C_{\text {out}} \times k^{2} \times E_{M A C},\\
        E_{S N N}=&(T \times D) \times f r \times O^{2} \times C_{in} \\&\times C_{out} \times k^{2} \times E_{A C},\nonumber
    \end{align}
    where $O$ is the output feature map size, $C_{in}$ and $C_{out}$ are the number of input and output channels, k is the kernel size, fr is the average spike firing rate, T is the number of time steps, and D is the virtual time step. Suppose the common direct training method is used, then $D=1$. Due to the homogeneity of convolution, the following BN and linear scaling transformations can be equivalently fused into the convolutional layer, and a bias is added during deployment. Therefore, when calculating the theoretical energy consumption, the energy consumption of the BN layer can be ignored. We use $E_{ANN}$ for the first convolutional layer of the encoder, and $E_{SNN}$ for the remaining parts. The fewer the spikes, the sparser the computation.
\section{SGReN Parameter Analysis}
To more clearly demonstrate the advantages of the SMAG module we proposed in terms of computational efficiency, this section will conduct a detailed comparison of the number of parameters and FLOPs between SMAG and the vanilla 1 $\times$ 1 convolution-based spiking MLP (SMLP). For a fair comparison, we assume that both have an input and output channel count of $C$. A typical SMLP structure includes two 1$\times$1 convolutional layers, a BN layer, and an activation function. Since the parameters of the BN layer and activation function are negligible, we only consider the convolutional layers in our calculation. Following the common setup for SNN-based Transformers, the expansion ratio of the SMLP is $R_{mlp}=4$, making the hidden layer channel count $4C$. The total parameter count $P_{SMLP}$ is easily derived as $8C^{2}$. The structure of SMAG is more complex, including partial convolution, 1$\times$1 convolution, and depthwise separable convolution. First, we use a 3$\times$3 kernel with a stride of 1 to perform partial convolution on $\frac{C}{4} $ of the channels:
\begin{align}
    P_{PConv}=3\times3 \times \frac{C}{4} \times  \frac{C}{4}, 
\end{align}
then we use a point-wise convolution to expand the channels to $4C$, followed by channel splitting. Features are extracted using depthwise convolution with kernel sizes of $3\times3$ and $7\times7$ respectively. The number of parameters of the deep convolution is:
\begin{align}
    P_{DWConv_{3\times3}}&=3\times3 \times C \times 1,\\
    P_{DWConv_{7\times7}}&=7\times7 \times C \times 1,
\end{align}
finally, a point-wise convolution is used to restore the channels to $C$. The total parameter count for the two point-wise convolutions is $P_{pwconv} = 6C^{2}$. Thus, the parameter count for SMAG is:
\begin{equation}
 \begin{aligned}
    P_{SMAG}&=P_{PConv}+ P_{DWConv}+P_{PWConv}\\
    &=6.5625C^{2}+58C. 
\end{aligned}   
\end{equation}

According to the theory of time complexity , when $C$ is large, the complexity is determined by the term with the highest power. Assuming the embedding dimension is consistent with QKFormer at 384 for the CIFAR dataset, the parameter count in the FFN stage can be reduced by approximately 16.08\%. The parameter count of SMAG is lower than that of the vanilla SMLP structure, mainly due to the high computational efficiency of depth-separable convolution in spatial feature extraction and the savings in computational load by partial convolution. Compared with SMLP, the SMAG reduces the number of parameters while achieving more diverse and numerous convolution, making feature extraction more refined and deeper.
\section{Frequency Analysis}
\begin{figure}[htb]
    \includegraphics[width=0.97\linewidth]{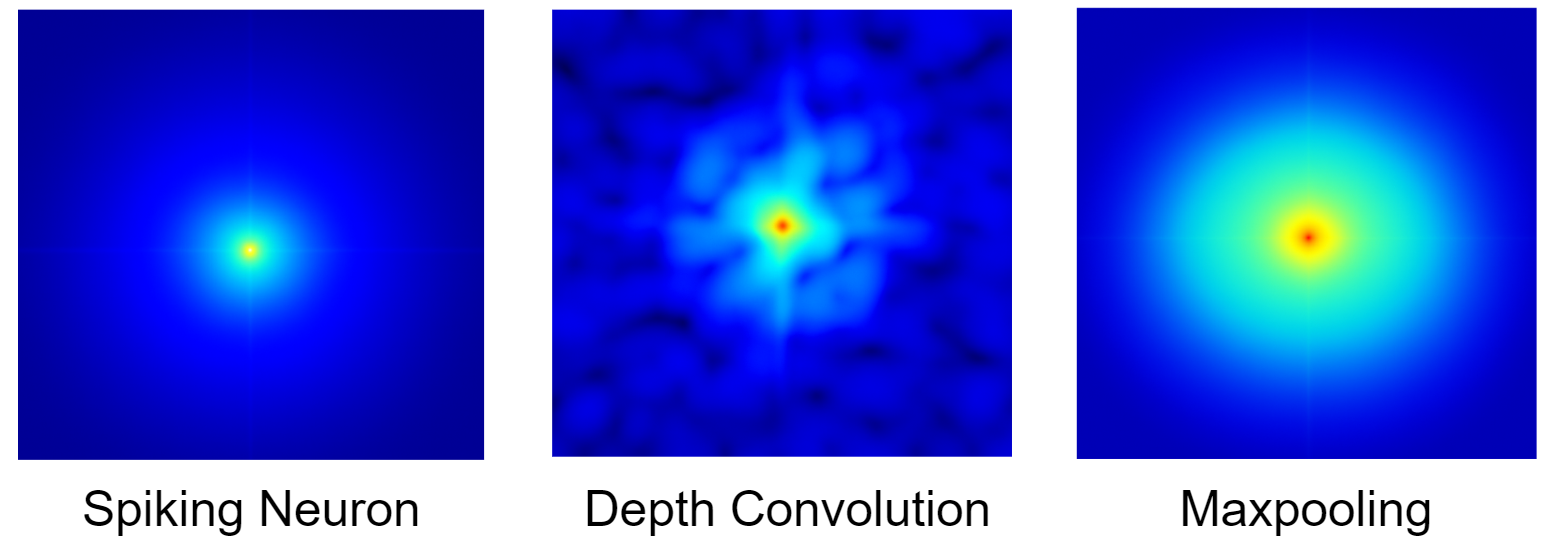}
    \caption{Fourier spectrum of Spiking Neurons, Spiking Depth-Wise Convolution and Spiking Max-Pooling.}
    \label{fourier}
\end{figure}
We conducted Fourier spectrum analysis on the feature maps after different operations, as shown in Fig. \ref{fourier}. It can be seen that the spectral energy of spiking neurons (LIFs) is highly concentrated in the central low-frequency region and rapidly decays as the frequency increases. This indicates that relying solely on spiking neurons can lead to significant loss of high-frequency information (such as edges and textures), thereby limiting the model's feature extraction ability at the shallow stage. In contrast, DWConv exhibits a more dispersed spectral response, especially retaining strong energy in the high-frequency regions far from the center. This means that DWConv can effectively capture and convey high-frequency details.

Based on the above observations, we strategically introduced DWConv and Max-Pooling in the Query/Key generation path of SAF Attention and in the Patch Embedding module. This design compensates for the intrinsic low-pass filtering characteristic of spiking neurons, actively preserving critical high-frequency visual cues at early network stages.

\section{Ablation Stduy}
\subsection{The effects of each component of SAF}
\begin{table}[htb]
\scalebox{0.97}{
\begin{tabular}{ccccc}
\toprule[1pt]
\multicolumn{1}{l}{Linear} & \multicolumn{1}{l}{DWConv} & \multicolumn{1}{l}{SGM} & \multicolumn{1}{l}{Params(M)} & \multicolumn{1}{l}{Accuracy(\%)} \\ \hline
\checkmark                          &                            &                         & 6.74                          & 82.95                             \\
\checkmark                          & \checkmark                          &                         & 6.76                          & 83.01                             \\
\checkmark                          &                            & \checkmark                       & 6.74                          & 83.10                             \\
\checkmark                          & \checkmark                          & \checkmark                       & 6.77                          & 83.38                             \\ \toprule[1pt]
\end{tabular}}
\caption{Ablation study of each component in the SAF module.}\label{ldsa}
\end{table}

To thoroughly investigate the individual contributions of each design component in SAF Attention, we conducted a progressive ablation study on CIFAR100. As shown in Table \ref{ldsa}, the baseline adopts conventional linear projections for Query/Key generation and discards the SGM module. Introducing depthwise separable convolution after Linear improves accuracy by 0.06\%, corroborating the efficacy of DWConv in preserving high-frequency local cues. Equipping the baseline with SGM alone yields a 0.15\% gain. When DWConv and SGM are integrated simultaneously, the model attains the peak accuracy of 83.38\%, surpassing the baseline by 0.43\%. This enhancement exceeds the naive summation of the two independent improvements, underscoring the greater significance of the SAF attention mechanism within the overall architecture.

\subsection{Different Kernel Size of DWConv on SMAG}
\begin{figure}[htb]
    \includegraphics[width=\linewidth]{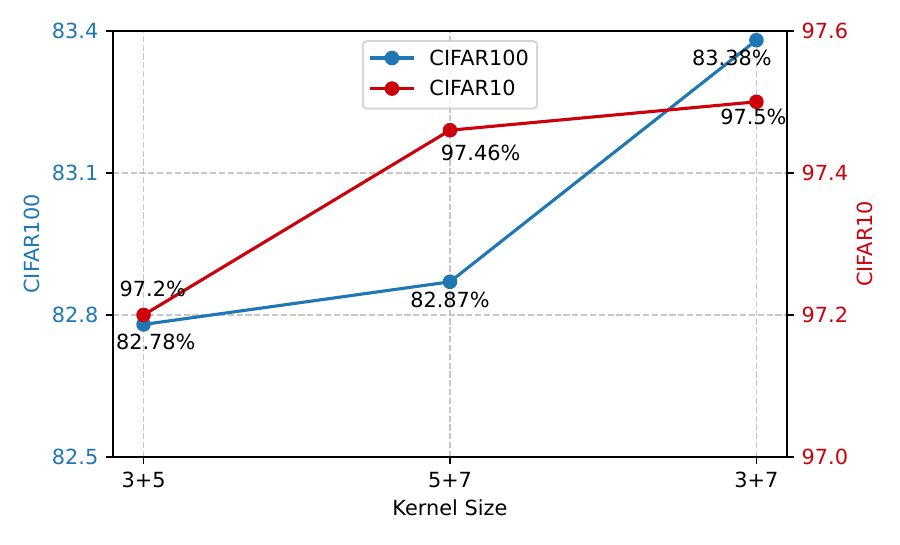}
    \caption{Ablation study on different kernel sizes of DWConv in the SMAG module.}
    \label{dw357}
\end{figure}
We conducted an ablation study on the optimal convolution kernel combination of multi-scale gated branches in SMAG. We compared three different configurations: $3\times3+5\times5$, $5\times5+7\times7$, and $3\times3+7\times7$. All experiments were conducted on the CIFAR10 and CIFAR100 datasets. As shown in Fig. \ref{dw357}, the combination of $3+7$ achieved the best performance on both datasets. The $3\times3$ kernel focuses on capturing fine local details, whears the $7\times7$ kernel is responsible for perceiving the broader regional context. This large span scale combination enables SMAG to take into account both local and global information simultaneously, generating the most discriminative and robust gating signals, thereby maximizing the effect of feature refinement.
\subsection{Generalizability and Modularity}
To demonstrate the transformative potential and generalizability of our SAF and SMAG modules, we integrated them into existing SOTA Spiking Transformers, SDT and QKFormer. As shown in Table~\ref{tab:modularity}, replacing the native attention in QKFormer with our SAF alone improves accuracy from 96.18\% to 96.51\%. These results unequivocally confirm SAF's role as a universal plug-and-play mechanism, enhancing various SNN backbones.

\begin{table}[htb]
\centering
\scalebox{0.85}{
\begin{tabular}{lcccc}
\toprule[1pt]
Model              & SAF & Acc(\%)            & SMAG & Acc(\%)            \\ \midrule
SDT                & \checkmark     & 96.10             &\checkmark      & 97.28              \\ 
QKFormer           & \checkmark     & 96.51          &\checkmark      & 97.25          \\ 
\textbf{SAFformer} &\checkmark     & \textbf{96.72} & \checkmark      & \textbf{97.31} \\ \bottomrule[1pt]
\end{tabular}
}
\caption{Applicability and modularity of SAF and SMAG on other SNN models (CIFAR-10).}
\label{tab:modularity}
\end{table}

\subsection{Effectiveness of SGM and Visualization}

The effectiveness of the SGM module is validated in Figure~\ref{saftime}. Compared to a static sparse version (Fixed Top-K with $K=0.6$), our SGM-enabled SAF achieves lower energy consumption across all scales. For the 27M model, SGM reduces energy by $\sim$27\% and accelerates training by 8\%. 

\begin{figure}[htb]
    \centering
    \includegraphics[width=0.95\linewidth]{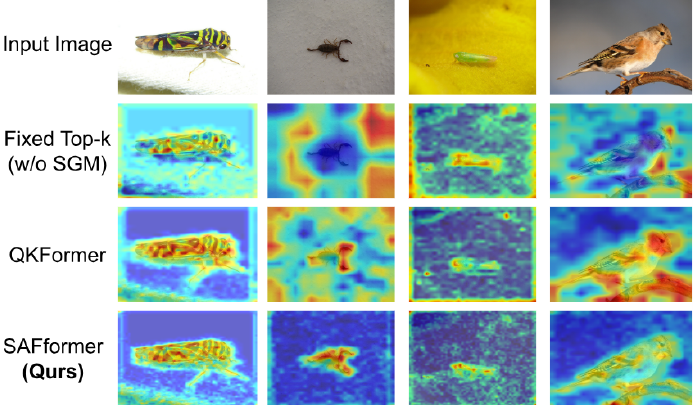}
    \caption{Visualization of attention heatmaps on ImageNet-1k. SAFformer precisely focuses on foreground objects and suppresses redundant background computation compared to Fixed Top-K and QKFormer.}
    \label{fig:visual_comp}
\end{figure}

As shown in Figure~\ref{fig:visual_comp}, SAFformer's attention is more refined. Quantitative statistics across 200 epochs on ImageNet-1K further reveal SGM's adaptive behavior: the average predicted $k$ value decreases from 0.5182 in Stage 1 to 0.4421 in Stage 2, reflecting SGM's ability to narrow attention as feature abstraction increases.

\subsection{Comparison with Mainstream ANN Models}

We further compare SAFformer with recent lightweight ANN models from 2024-2025. As shown in Table~\ref{tab:ann_comp}, our model achieves a superior accuracy-efficiency tradeoff. While models like CrossFormer++-T and SpectFormer-S achieve high accuracy, SAFformer significantly reduces energy consumption to only 5.88mJ.

\begin{table}[htb]
\centering
\scalebox{0.85}{
\begin{tabular}{lccc}
\toprule[1pt]
Model              & Param(M) & Power(mJ) & Acc(\%)  \\ \midrule
CrossFormer++-T   & 27.80                        & 13.34                         & 81.50                        \\ 
SpectFormer-S       & 32.56                        & 30.36                         & \textbf{81.70}                        \\ 
\textbf{SAFformer} & 26.58                        & \textbf{5.88}                 & 80.44$\pm$0.24 \\ \bottomrule[1pt]
\end{tabular}}
\caption{Comparison with mainstream ANN methods on ImageNet.}
\label{tab:ann_comp}
\end{table}

\section{Further Discussion on Proposition 2}

To gain a deeper understanding of the theoretical implications of Proposition~2, we contrast our \textbf{Active Predictive Filtering} paradigm with the \textbf{Passive Reactive Sparsity} inherent in conventional Spiking Transformers. While both lead to sparse spike activity, the underlying computational logic and information-theoretic motivations are fundamentally distinct.

\textbf{Externalization of Filtering Decisions.} 
In standard SNNs (e.g., Spikformer or standard SSA), sparsity is a passive byproduct of local signal strength. A neuron fires only when its accumulated membrane potential exceeds a fixed threshold $V_{th}$. This is a ``compute-first, filter-later'' process where the filtering decision is internal to the data path. In contrast, as formalized in Proposition~2, SAF decouples the saliency evaluation from the information filtering. The introduction of a global selection function $\Psi(X, k)$, realized by the SGM, represents an externalized decision-making mechanism. The SGM acts as a ``meta-controller'' that observes the global context $X$ to predict the optimal computation graph before heavy neuronal integration occurs. This mirrors the top-down modulation in the biological visual system, where higher-level expectations proactively prime specific neural circuits.

\textbf{Topological Reconfiguration of Information Flow.} 
Proposition~2 defines the SAF computation on a dynamically selected sparse subgraph $G'$. This is a crucial distinction from fixed-topology sparse Transformers. In our framework, the graph's topology—the set of active vertices $V_{\text{active}}$—is not static but is proactively reconfigured for every input. For ``simple'' images with redundant backgrounds, the SGM predicts a smaller $k$, effectively pruning the majority of the graph and focusing computational resources exclusively on the object of interest. For ``complex'' scenes, $k$ is adaptively increased. This content-dependent graph reconstruction allows SAFformer to maintain a high information bottleneck for noise while ensuring sufficient bandwidth for salient ``prediction errors.''

\textbf{Synergy between Contextualization and Saliency.} 
The proof of Proposition~2 relies on the unary saliency function $S(v_i)$. Crucially, $v_i$ is not a raw token but a contextualized representation enhanced by depthwise convolutions. This architectural choice ensures that the ``filtering'' is guided by high-level spatial information (e.g., edges, textures) rather than just pixel-wise intensity. By proving that SAF operates as a node-wise filter on a predicted subgraph, we establish a theoretical foundation for its superior energy efficiency: computation is only expended on nodes that have been ``authorized'' by the SGM's global prediction.

In summary, Proposition~2 clarifies that SAF's sparsity is a deliberate, learned strategy derived from context, rather than a reactive consequence of weak activations. This paradigm shift from ``passive thresholding'' to ``active predictive filtering'' is the core driver behind SAFformer's performance and extreme energy efficiency.